\documentclass{article} 
\usepackage{iclr2025_conference,times}
\usepackage{graphicx}
\usepackage{subfigure}

\usepackage{amsmath,amsfonts,bm}

\def\eqref#1{equation~\ref{#1}}

\def\1{\bm{1}}

\DeclareMathAlphabet{\mathsfit}{\encodingdefault}{\sfdefault}{m}{sl}
\SetMathAlphabet{\mathsfit}{bold}{\encodingdefault}{\sfdefault}{bx}{n}

\usepackage{hyperref}
\usepackage{url}

\usepackage{amsmath}
\usepackage{amssymb}
\usepackage{mathtools}
\usepackage{amsthm}

\usepackage{amsmath}
\usepackage{algorithm}
\usepackage{algpseudocode}

\usepackage{CJKutf8}

\usepackage{booktabs}
\usepackage{longtable}
\usepackage{caption}
\usepackage{array}

\theoremstyle{plain}
\newtheorem{theorem}{Theorem}[section]

\theoremstyle{definition}

\newtheorem{assumption}[theorem]{Assumption}
\theoremstyle{remark}

\title{AdaMemento: Adaptive Memory-Assisted Policy Optimization for Reinforcement Learning}

\author{Renye Yan\textsuperscript{\rm 1}\footnotemark[1],  Yaozhong Gan\thanks{Equal contribution.}, You Wu, Junliang Xing\textsuperscript{\rm 2}, Ling Liangn\textsuperscript{\rm 1}, Yeshang Zhu, Yimao Cai\textsuperscript{\rm 1}\\
\textsuperscript{\rm 1}Peking University, \textsuperscript{\rm 2} Tsinghua University\\
\texttt{victory@stu.pku.edu.cn, yzgancn@163.com}
}

\newcommand{\ourmethod}{AdaMemento}

\iclrfinalcopy 
\begin{document}

\begin{CJK*}{UTF8}{gbsn}

\maketitle

\begin{abstract}
In sparse reward scenarios of reinforcement learning (RL), the memory mechanism provides promising shortcuts to policy optimization by reflecting on past experiences like humans. However, current memory-based RL methods simply store and reuse high-value policies, lacking a deeper refining and filtering of diverse past experiences and hence limiting the capability of memory. In this paper, we propose \ourmethod{}, an adaptive memory-enhanced RL framework.
Instead of just memorizing positive past experiences, we design a memory-reflection module that exploits both positive and negative experiences by learning to predict known local optimal policies based on real-time states. To effectively gather informative trajectories for the memory, we further introduce a fine-grained intrinsic motivation paradigm, where nuances in similar states can be precisely distinguished to guide exploration. The exploitation of past experiences and exploration of new policies are then adaptively coordinated by ensemble learning to approach the global optimum. Furthermore, we theoretically prove the superiority of our new intrinsic motivation and ensemble mechanism. From 59 quantitative and visualization experiments, we confirm that \ourmethod{} can distinguish subtle states for better exploration and effectively exploiting past experiences in memory, achieving significant improvement over previous methods.

\end{abstract}

\section{Introduction}

In reinforcement learning (RL), policy optimization directly updates the policy function according to reward signals and hence efficiently converges to the optimum in simple tasks. However, in sparse reward environments, policy updates become unstable and ineffective due to insufficient feedback~\citep{bellemare2016unifying,liang2018memory}.
This significantly increases the difficulty of learning effective long-horizon policies.

Memory offers a promising solution to the sparse reward problem, as humans can effectively learn from past experiences to avoid repeating mistakes in similar scenarios~\citep{liu2021human,bransford1972contextual,andrychowicz2017hindsight}.
Through memory, agents can utilize prior successful experiences to refine their policies in complex environments, hence reducing the reliance on dense reward feedback and improving both learning efficiency and policy stability~\citep{pathak2017curiosity}.

Existing memory-based RL methods can be roughly categorized into two classes.
Storage-based methods like NGU~\citep{badia2020never} maintain a memory buffer of visited states and avoid re-visitation of similar states. 
This kind of approach essentially treats memory as a state counter, without leveraging past experiences to predict or guide future policies.
Experience replay methods, such as self-imitation learning (SIL)~\citep{oh2018self}, aim to reuse past high-value experiences to optimize the policy by repeatedly imitating state-action pairs with a higher value.
\citet{guo2020memory} further extend SIL to replicate historical trajectories and guide the agent to explore on top of these paths. While existing experience replay methods have shown promising capabilities, they overly rely on past experiences and pay less attention to exploration, which may lead to premature policies. Moreover, these methods neglect the importance of failed experiences, despite the fact that failures often lie along the high-value trajectories. Therefore, ideal memory-based RL, like humans, should effectively generalize from both successful and failed experiences, as well as balancing the exploitation of past experiences and the exploration of new states.

To address the above challenges, in this paper, we propose \ourmethod{}, a novel RL framework inspired by the memory reflection process of humans. \ourmethod{} consists of a memory-reflection module for exploitation and a coarse-fine distinction module for exploration. The former focuses on refining the generality from past experiences in memory, while the latter continues to discover new policies in a fine-grained way and update the memory. The two modules are trained separately and then adaptively combined through ensemble learning. Altogether, the exploitation of memory promotes deeper exploration, which in turn provides high-quality experiences for effective exploitation.

Specifically, the memory-reflection module maintains memory buffers of successful and failed trajectories to train two networks to predict and evaluate actions based on real-time states. Firstly, the prediction network summarizes past trajectories into a local shortest-path policy. Then the reflection network receives both successful and failed experiences for memory reflection training, outputting confidence of the predicted policy. Collaboratively, the memory prediction and reflection learn the mapping from states to the optimal policy, enabling the agent to follow the current shortest path to the endpoint of the previous trajectory and continue exploring.

\begin{figure*}[t]
    \centering
    \includegraphics[width=1.0\linewidth]{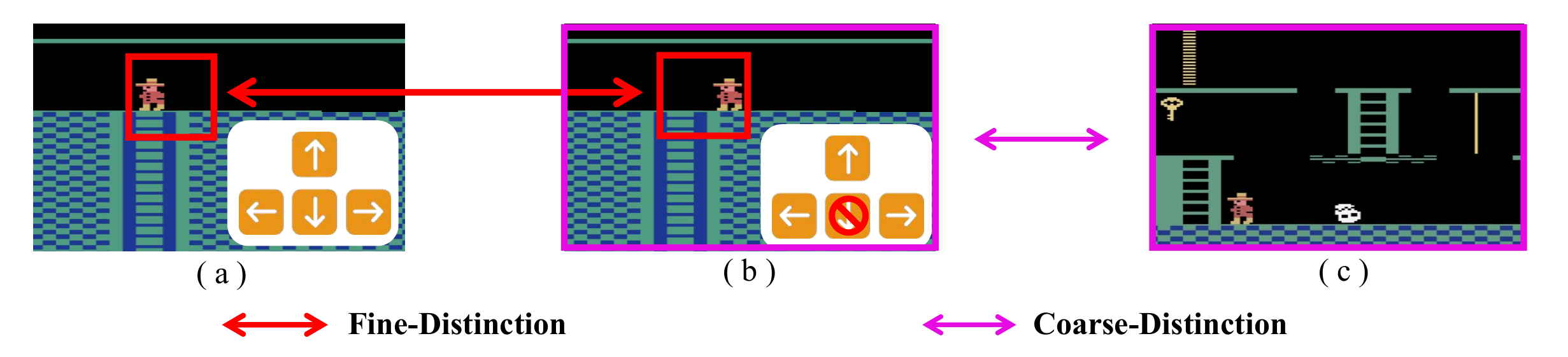}
    \caption{Different granularity of state discrimination. (a) versus (b) represents fine-grained distinction, where the state images look similar but are of completely different importance, which is not well addressed in previous research.}
    \label{qifen}
\end{figure*}

Moreover, we argue that the effectiveness of memory depends heavily on the quality of data gathered during exploration. Therefore, instead of solely focusing on the memory mechanism, \ourmethod{} also improves the exploration policy and emphasizes on its collaboration with memory.
Existing exploration-centric methods formulate intrinsic rewards based on states' visual novelty~\citep{burda2018exploration,pathak2017curiosity,guo2022byol}. However, the novelty based on the whole state is too coarse to precisely distinguish the potential importance differences of visually similar states, as illustrated in Figure~\ref{qifen}.
We propose a coarse-fine distinction module, which not only discovers novel states through reconstruction error of raw state images but also enables fine-grained discrimination of similar states via sparsity of latent representations. The fine-grained incentive avoids confusion between similar states and increases the likelihood of discovering potentially better policies.

\begin{figure*}[t]
    \centering
    \includegraphics[width=1.0\linewidth]{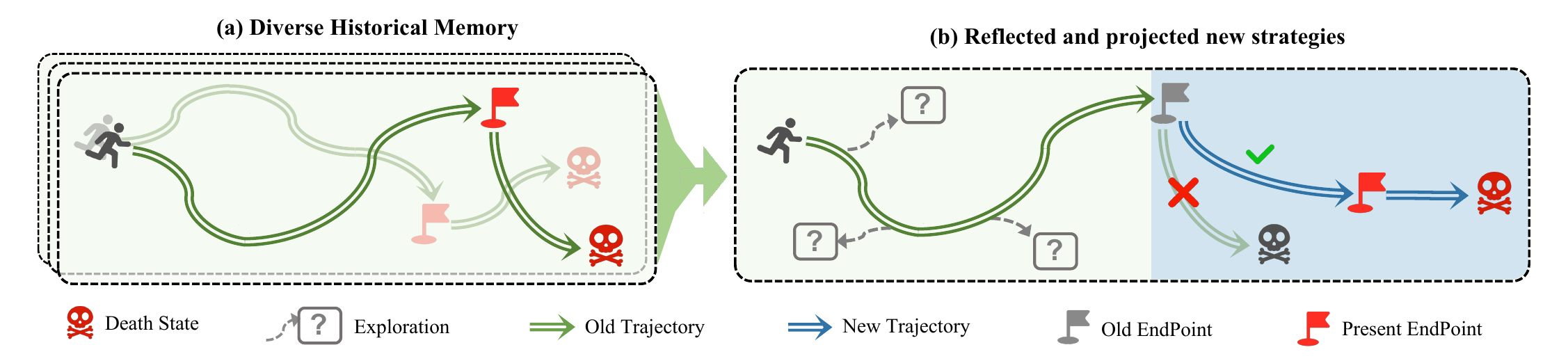}
    \caption{The figure shows that the left side represents past experience trajectories stored in the memory buffer. AdaMemento learns to avoid danger and continues updating the current optimal strategy by synthesizing and reflecting on the commonalities in these trajectories. The updated strategy is illustrated on the right side.}
    \label{tra}
\end{figure*}

In summary, \ourmethod{}'s memory mechanism considers both successful and failed policies in that it draws the shortest path from positive experiences and reflects on how to avoid risks from negative ones, as illustrated in Figure~\ref{tra}. Thus, our exploitation of memory differs from simple experience replay or imitation (e.g., SIL, DTSIL). By incorporating fine-grained state distinction into intrinsic rewards, we establish a new intrinsic reward paradigm and theoretically prove its validity. Finally, through ensemble learning, \ourmethod{} achieves an adaptive balance between exploration and exploitation by adaptively choosing actions from the memory-based prediction and the enhanced exploration based on evaluated confidence after reflection. We theoretically prove that the coordinated policy outperforms the original one.

To thoroughly validate the effectiveness and generalization ability of \ourmethod{}, we tested its performance across 56 Atari and MuJoCo environments. Experimental results demonstrate that \ourmethod{} performs well and stably generalizes to a broad range of environments, from discrete to continuous and from sparse reward to dense reward settings. Especially in the most challenging Montezuma's Revenge, \ourmethod{} achieves more than 15 times gain in total rewards. In addition, to clearly illustrate the effectiveness of each module on exploration and exploitation capabilities, we conduct visualization experiments. The results show that the ensemble of memory reflection helps the agent to reach the goal with significantly fewer steps in hard environments, and the coarse-fine distinction enables a larger range of and finer-grained exploration even when rewards are extremely sparse. Besides, it is important to note that the major hyper-parameters of \ourmethod{} are kept the same across all experimental environments, confirming the advantage in robustness of our method.
In summary, this paper presents the following three contributions:
\begin{itemize}
    \item We propose a memory-reflection mechanism that embeds both successful and failed experiences into network weights, which, unlike the imitation in existing memory methods, enables more accurate prediction of trajectories and confidence levels.
    \item To supply memory with high-quality experience data, we propose a new paradigm of intrinsic motivation based on reconstruction errors of state images and sparsity of latent encodings, for which we provide theoretical proof of no deviation from the optimal policy.
    \item We leverage ensemble learning to adaptively coordinate between the exploitation of memory and the exploration of new policies, which makes \ourmethod{} flexible to integrate with existing methods and boosts their performance with a theoretical guarantee.
\end{itemize}

\section{Related work}
\subsection{Memory-based RL}

Current memory-based RL methods can be divided into two main categories, i.e., storage-based methods and experience replay. Storage-based methods retain past states and manage the memory buffer using clustering, as seen in works like NGU~\citep{badia2020never} and BT-NGU~\citep{campos2021beyond}. These methods compare the current state with the stored cluster centers to help the agent discover novel states and enhance exploration efficiency. However, these approaches are essentially state counting via a memory buffer rather than fully utilizing the knowledge of past experiences. 
Experience replay methods, such as SIL~\citep{oh2018self} and DTSIL~\citep{guo2020memory}, accelerate policy convergence by replaying past advantageous experiences. However, experience replay often overlooks the importance of exploring new policies and reflecting on failed experiences, which can lead to suboptimal policies. 
Moreover, existing memory-based RL methods often manage the memory buffer through clustering, which can lead to incorrectly clustering similar states with different values in high-dimensional state spaces. This neglects critical nuances, potentially missing opportunities for policy optimization.

\subsection{Intrinsic motivation}

Intrinsic motivation models based on curiosity to guide exploration have gained attention for addressing sparse and delayed rewards. These methods can be broadly categorized into two types: count-based and prediction error-based. Count-based methods~\citep{bellemare2016unifying} track state visitation frequency and provide higher rewards for less-explored states. While simple and intuitive, they suffer from significant storage pressure in complex environments due to the large state space. In contrast, prediction error-based methods~\citep{burda2018exploration,pathak2017curiosity} evaluate state density using neural networks and reward more on the low-density states, without requiring additional storage, offering a notable improvement in constructing intrinsic motivation compared to count-based approaches.

Current intrinsic motivation models essentially rely on state novelty as the measure of intrinsic rewards. While these methods perform well in early exploration, existing metrics of state novelty are insufficient. As the training progresses, the marginal benefit of intrinsic rewards diminishes once most states have been explored, leading to a significant drop in exploration efficiency and to the ``aimless exploration dilemma"~\citep{barto2013novelty}. A finer-grained metric for that can distinguish the nuances of similar states is needed for better intrinsic motivation.
\begin{figure*}[t]
    \centering
    \includegraphics[width=1.0\linewidth]{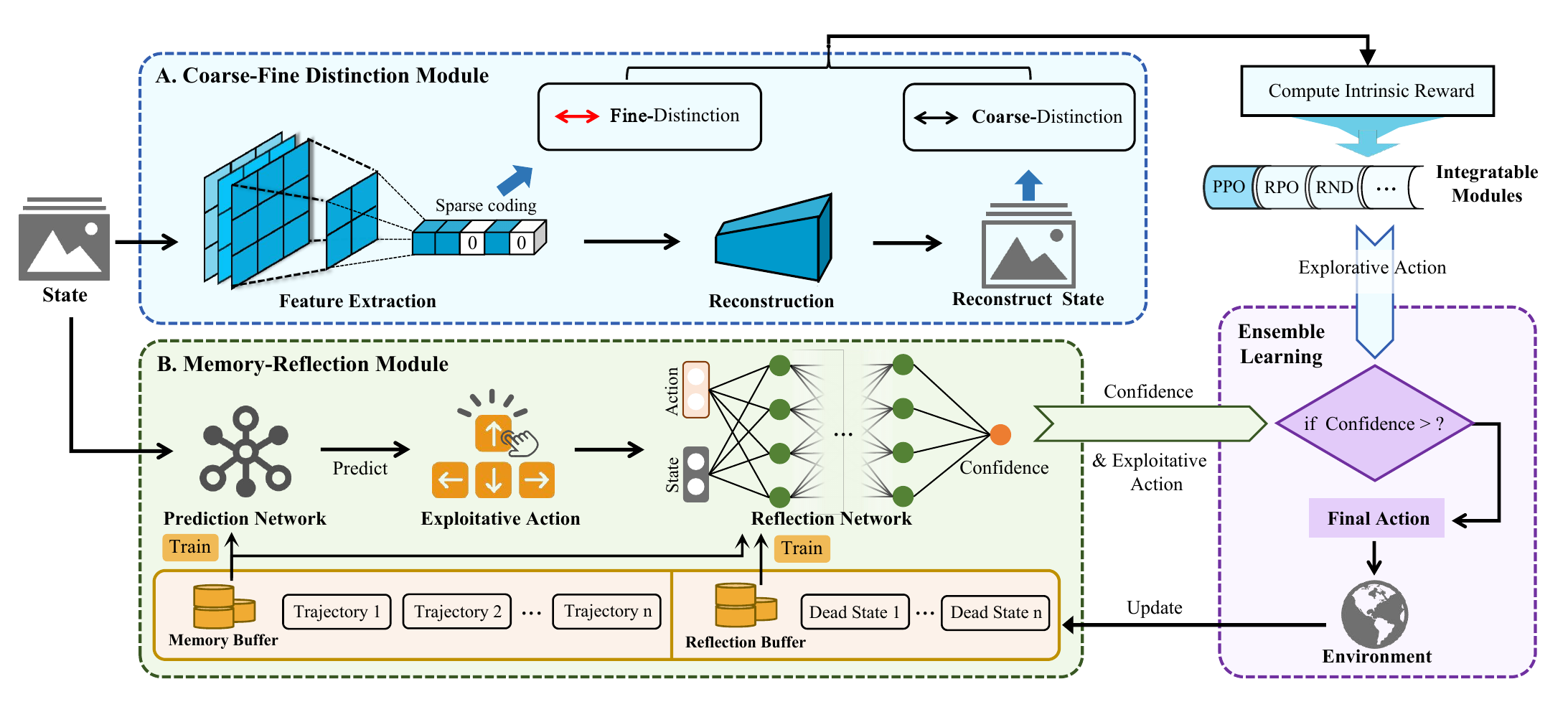}
    \caption{\textbf{\ourmethod{}'s framework.} We evaluate each sub-module in (a) and parameters in (b) and (c).}
    \label{framework}
\end{figure*}

\section{Method}
\label{method}

In RL, refining and extracting common patterns from diverse memory experiences to guide subsequent exploration is more efficient than starting from scratch.
Effective utilization of memory requires not only reflecting on and integrating diverse past experiences but also incorporating sufficient exploration to approach a globally optimal policy. Based on this motivation, we propose the \ourmethod{}, an innovative memory-guided RL framework. As illustrated in Figure~\ref{framework}, \ourmethod{} enhances the policy with a memory-reflection module for exploitation and a coarse-fine distinction module for exploration, and carefully balances the two mechanisms.
In Section~\ref{method_1}, we describe the design of the memory-reflection mechanism. In Section~\ref{method_2}, we introduce the coarse-fine distinction mechanism and prove the rationality of the new paradigm of intrinsic rewards; in Section~\ref{method_3}, we employ ensemble learning to achieve adaptive coordination between exploration and exploitation, the effectiveness of which is theoretically guaranteed.

\subsection{Memory-reflection module}
\label{method_1}

Unlike previous approaches that rely solely on successful experiences, an effective memory mechanism should extract valuable insights from successful experiences and learn to avoid risks from failures, which means the agent should possess the ability to reflect on memory. Therefore, we introduce a memory-reflection module that incorporates both positive and negative experiences to improve the utilization of memory, as illustrated in Figure~\ref{framework}(A). During training, we maintain two fix-sized memory buffers of history trajectories and use them to train a prediction network and a reflection network.

\subsubsection{Prediction network} 

\begin{figure*}[t]
    \centering
    \includegraphics[width=1.0\linewidth]{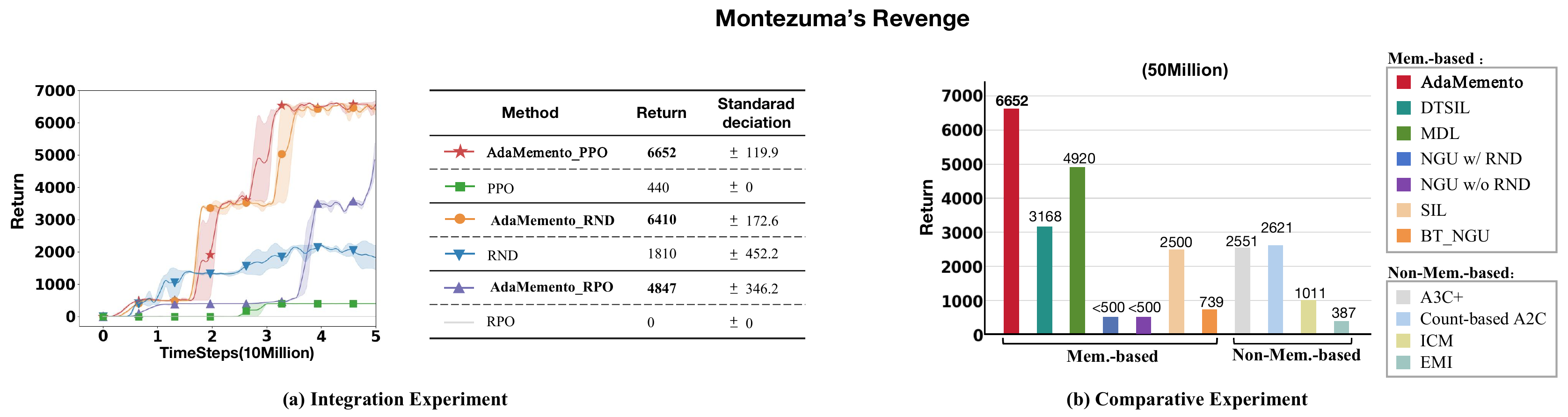}
    \caption{\textbf{Comparison in Montezuma's Revenge Environment. }(a) illustrates a comparison between baseline methods before and after integration with our \ourmethod{}; (b) presents a performance comparison to other advanced baseline models.}
    \label{montezumamain}
\end{figure*}
In reality, common states in different trajectories often form the shortest path to the goal, making extracting and summarizing these shared experiences more critical than simply mimicking a single trajectory.
Therefore, unlike traditional memory methods that simply replicate or imitate specific trajectories, we design the prediction network to extract and summarize common experiences. It independently trains on state-action pairs selected from a memory buffer of past advantageous trajectories. The benefit of this approach is that the prediction network can identify and embed the common patterns of advantageous experiences into the network weights, thereby learning to capture the underlying policy that links states to actions rather than simply repeating past actions, which can be applied across various environments and conditions.

Given a pair of state $s$ and the actual action $a$ from the advantageous memory buffer, we align the output action $\hat{a}$ of the prediction network with the goal of policy optimization, using the cross-entropy loss:
\begin{equation}
loss(\pi_{\theta}) = - \sum_{(s,a)\in M}\pi(a|s)\log \pi_{\theta}(\hat{a}|s),
\end{equation}
where $\pi_\theta$ and $\pi$ represent the policy from the prediction network and the policy drawn from the memory buffer of advantageous trajectories, respectively.

After training, the prediction network approximates the current local optimal policy from the consensus of past experiences to help the agent predict action sequences. This mechanism enables the agent to quickly leverage the shared patterns of advantageous trajectories from history to guide the agent to follow the shortest path and reach the previous endpoint to continue exploring, which improves training efficiency and accelerates the convergence to the globally optimal policy.

\subsubsection{Reflection network}

The reflection network is responsible for assessing the memory-based policy from the prediction network and outputting a corresponding confidence level ($C_{\omega}$) indicating the quality of the action $\hat{a}$. If $\hat{a}$ tends to lead the agent to failure, the reflection confidence will be decreased; conversely, if $\hat{a}$ contributes to the agent's success, the reflection confidence is increased. The loss function is defined as follows:
\begin{equation}
    loss_C(\omega)=\!\sum_{(s,a)\in M}\!\!\|C_{\omega}(s,a)-1\|^2+\!\sum_{(s,a)\in R}\!\!\|C_{\omega}(s,a)-0\|^2\!,
\end{equation}
where $M$ and $R$ represent actions from two memory buffers, respectively, one containing successful experiences for memorization (referred to as $M$-buffer) and the other containing failed experiences for reflection (referred to as $R$-buffer). Details of the two memory buffers are described in Appendix~\ref{supplementary_buffers}.

Note that the above prediction network is trained only using the $M$-buffer while the reflection network is trained on both the $M$-buffer and the $R$-buffer.
We force the confidence level of the action from the $M$-buffer to be 1 and the action from the $R$-buffer to be 0.
This mechanism enhances the robustness of the overall decision-making system by offering dynamic feedback and optimization of prediction accuracy.

\subsection{Coarse-fine distinction module}
\label{method_2}
\begin{figure*}
    \centering
    \includegraphics[width=1.0\linewidth]{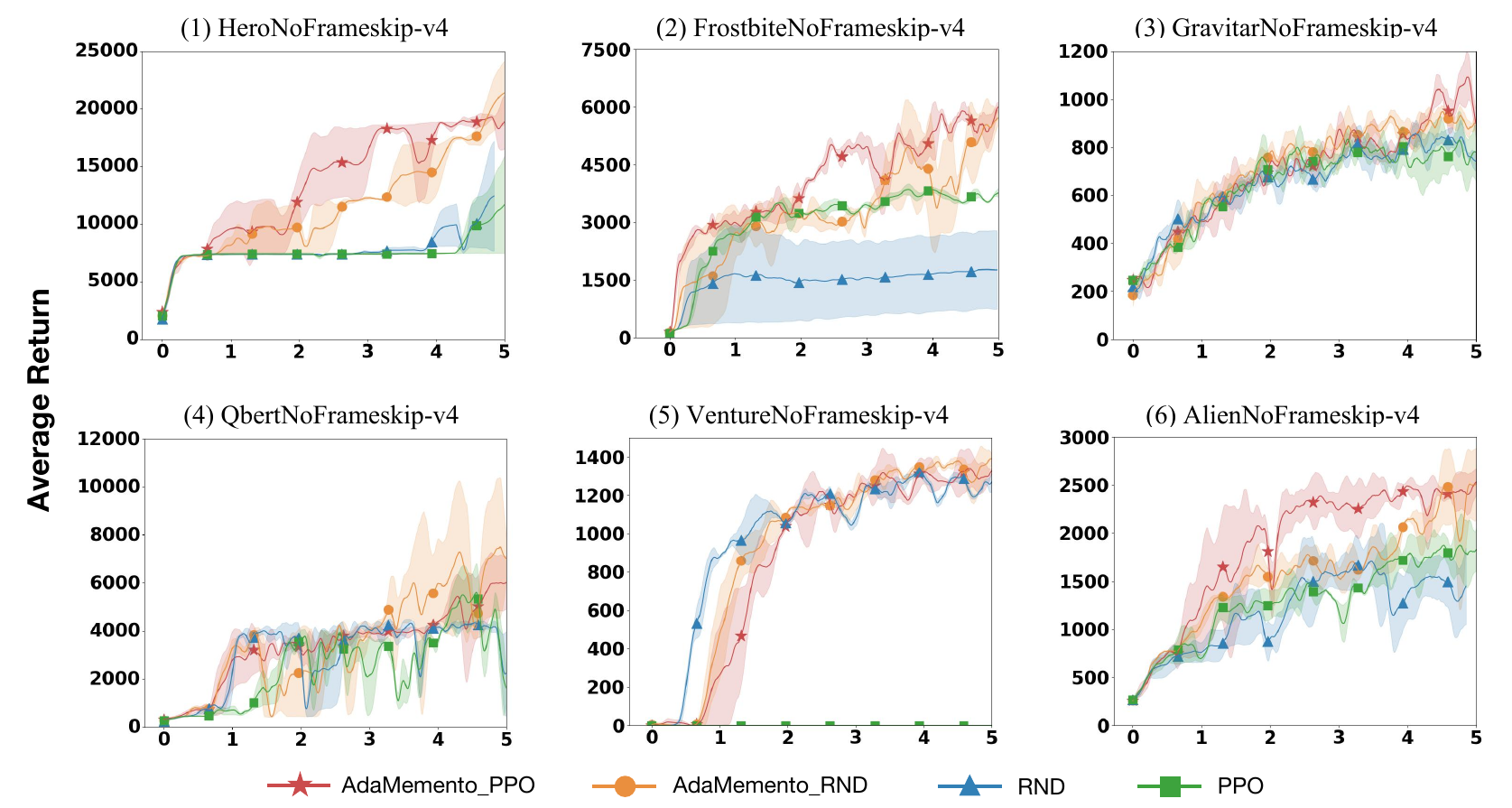}
    \caption{\textbf{Generalization experiments in discrete-space environments (Atari).} The x-axis represents timesteps in 10 million.}
    \label{atariothers}
\end{figure*}
Memory reflection requires diverse historical experiences to progressively approach the global optimum, which depends on high-quality exploration, particularly in complex tasks with sparse rewards. To address this, we propose an innovative coarse-fine distinction module, where we establish a new intrinsic motivation paradigm and introduce a fine-grained distinction mechanism. As illustrated in Figure~\ref{framework}(B), this paradigm integrates both state novelty and fine-grained distinctions between similar states while unifying the training of exploration policy networks with intrinsic motivation metrics. We also provide theoretical proof for its retaining of the optimal policy. Therefore, it not only enhances the generalizability of fine-grained exploration. More importantly, it provides high-quality exploration data to support the exploitation of memory proposed in this work.

\subsubsection{Intrinsic reward design}

As illustrated in Figure~\ref{framework}(A), our intrinsic reward design is based on a two-level distinction scheme responsible for coarse state novelty and fine-grained distinctions between similar states, denoted as C-Discriminator and F-Discriminator respectively.
Given a raw image of a state as the input \( s_t \), the C-Discriminator learns the feature representations \( f_\theta(s_t) \) and reconstructs the image $\hat{s}_t = g_{\phi}(f_\theta(s_t))$. Then we use the reconstruction error between $\hat{s}_t$ and \( s_t \) as the coarse-grained measurement to identify novel images from the frequently trained images. For example, the states that have been explored many times can be distinguished from the states similar to those in previous scenarios due to their different fitting degree in the reconstruction network.
Therefore, the reconstruction error can be used to guide the agent to explore unknown or infrequent states.

The C-Discriminator is useful for novel state exploration, but reconstruction error cannot distinguish visually similar states, which is necessary for challenging environments with subtle feedback. For instance, consecutive images may have similar reconstruction errors from the network, but the importance they represent could be entirely different, e.g., an image close to the reward versus its nearby image that traps the agent (Figure~\ref{qifen}(a) versus \ref{qifen}(b)). Therefore, we further propose an innovative F-Discriminator to support the fine-grained distinction of states. The F-Discriminator poses a sparse constraint to the feature vector \( f_\theta(s_t) \) with a $\ell_1$-norm regularization.
By analyzing the sparsity of $f_\theta(s_t)$, we can precisely distinguish between trained states, states similar to trained ones, and entirely new states.

We unify the exploration policy's network training loss with the coarse-fine intrinsic motivation measure, formulating our new paradigm as below:
\begin{equation}
\label{unequ}
R_I =loss_{f,g}(\theta, \phi) = \frac{1}{2} \left\| s_t - g_{\phi}(f_\theta(s_t)) \right\|_2^2 + \lambda \left\| f_\theta(s_t) \right\|_1,
\end{equation}
where $f_\theta$ and $g_{\phi}$ respectively denote the feature extraction network (encoder) and reconstruction network (decoder).

This paradigm innovatively distinguishes the importance of closely related similar images, thereby improving the accuracy of intrinsic rewards and enabling more precise allocation of exploration effort. Furthermore, the unified paradigm integrates network loss and intrinsic reward evaluation, avoiding the need for manual design across different tasks and enhancing its generalization ability for multi-task adaptation. The coarse-fine motivation not only achieves more precise state exploration but also provides more accurate learning samples for memory learning. Subsequent visualization experiments further validate these findings.

\subsubsection{Proof of the retaining of policy optimality}
In RL, ensuring the reliability of intrinsic rewards is crucial, especially in environments with sparse rewards. For our \ourmethod{} design, introducing the coarse-fine intrinsic exploration reward must not skew or misguide the learning agent away from its optimal policy. Here, we aim to ascertain the impact of our intrinsic exploration reward on the policy's optimality. We start by introducing an assumption.

\begin{assumption}\label{assu1}
After $k$ updates, we assume $0\leq \beta R_I \leq (1-\gamma)\min_{s}[Q^*(s, a^*)-Q^*(s,a^{sub})]$ holds, where $Q^*$ is the optimal $Q$ function, $a^*$ and $a^{sub}$ are optimal and suboptimal actions under the state $s$, respectively.
\end{assumption}

Assumption~\ref{assu1} is reasonable as the networks are trained to minimize the losses, and hence, the accompanying intrinsic rewards will be smaller enough after certain steps of training. Under this trivial assumption, we can prove that our new intrinsic motivation paradigm will retain the optimality of policy.

\begin{theorem}\label{the_same_ap1}
    After $k$ updates of the coarse-fine distinction network, under Assumption \ref{assu1}, the optimal action remains the same after adding the intrinsic rewards. That is, for any state $s$, we have
    \begin{equation}
        \arg\max_a Q^*(s,a)=\arg\max_a Q^*_1(s,a),
    \end{equation} 
    where $Q^*_1$ is the optimal $Q$ function after adding the intrinsic rewards.
\end{theorem}

The proof is shown in Appendix~\ref{appendix_proof1}. Theorem~\ref{the_same_ap1} reveals the rationale behind the intrinsic reward design based on loss functions, which has not been mentioned in existing work like RND. It also points out a new direction for future intrinsic reward design.

\begin{figure*}[t]
    \centering
    \includegraphics[width=1.0\linewidth]{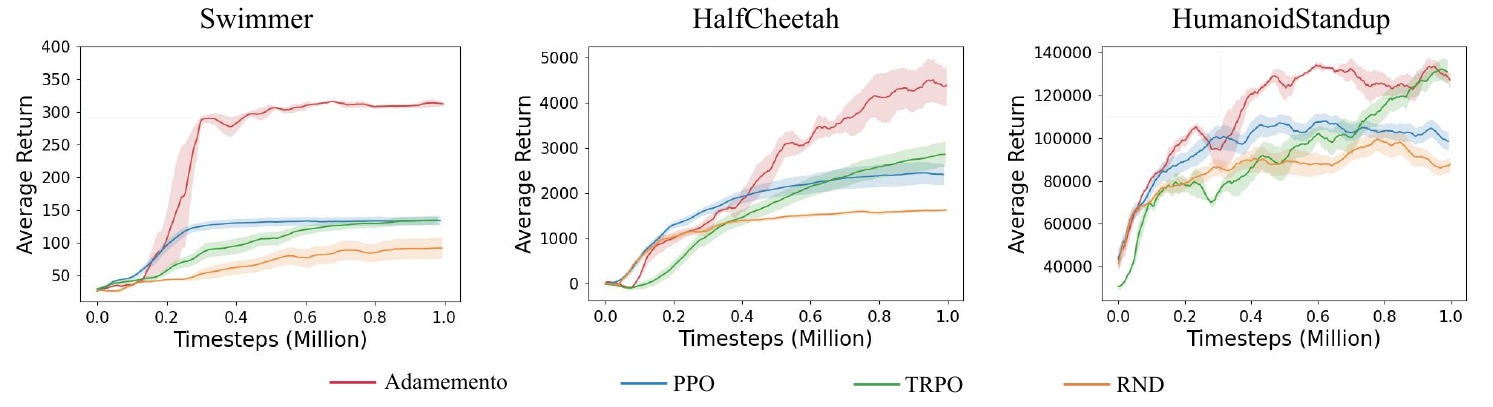}
    \caption{\textbf{Generalization experiments in MuJuCo}, showing the generalizable performance of \ourmethod{} in different types of \textbf{continuous space tasks}. The x-axis represents timesteps in millions.}
    \label{mujucoothers}
\end{figure*}
\subsection{Ensemble learning for exploration-exploitation balance}
\label{method_3}
The memory-reflection module and the coarse-fine distinction guide exploitation and exploration respectively. It is critical for the final policy to balance between the two branches of policy. Hence, we leverage ensemble learning to adaptively choose actions.
Denote the actions from the memory reflection as $a_{mem}$ and actions given by the original RL algorithm with our intrinsic rewards as $a_{ori}$. Given a policy \( \pi \), e.g., PPO, we establish its greedy counterpart denoted as \( \pi_\theta \)
and define a combination function:
\begin{equation}
\label{pi}
   \pi_{new} = (1-I_{\{C_{\omega}(s, a)\geq \kappa\}}) \pi + I_{\{C_{\omega}(s, a)\geq \kappa\}} \pi_\theta,
\end{equation}
where \(\kappa\) is the confidence threshold, and \(I_{\{C_{\omega}(s, a)\geq \kappa\}}=1\) when the prediction of memory reflection network \(C_{\omega}(s, a)\geq \kappa\) is larger than the threshold \(\kappa\). It is important to note that we balance exploration and exploitation directly on the policy level instead of rewards and that \( I_{\{C_{\omega}(s, a)\geq \kappa\}} \) is entirely based on the network's real-time output rather than being defined by human experience.

\begin{figure*}[t]
    \centering
    \includegraphics[width=1.0\linewidth]{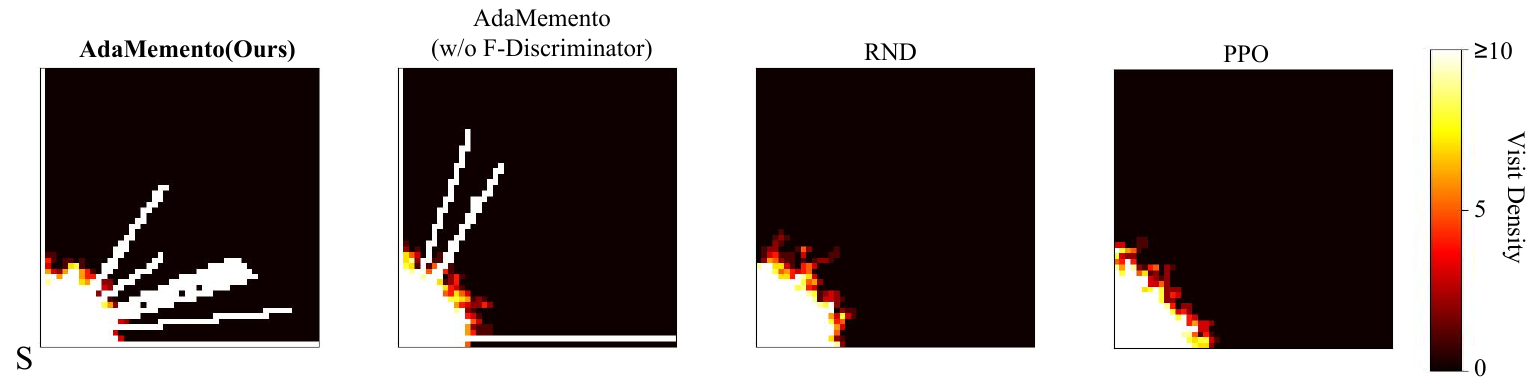}
    \caption{\textbf{Exploration Visualization in Dark Chamber} in 50k steps. The exploration radius of \ourmethod{} with the F-Discriminator is the largest, leading to the largest number of valid paths.}
    \label{darkcharmber}
\end{figure*}
\subsubsection{Proof of memory effectiveness}
Through ensemble learning, AdaMemento combines exploration and exploitation and adaptively selects between memory prediction or exploration based on the confidence provided by the Reflection Network. our ensemble learning achieves adaptive coordination between exploration and exploitation.
To underscore the efficacy of this approach, we present Theorem~\ref{the_same_ap2} that highlights the policy value of \( \pi_{\text{new}} \) is at least as good as that of \( \pi \).

\begin{theorem}\label{the_same_ap2}
    According to Eqn. (\ref{pi}), if $\pi_\theta$ is a greedy policy for $\pi$, that is, for any state $s$, it have $\pi_\theta(a^*|s)=1$, where $a^*=\arg\max_{a}Q^{\pi}(s,a)$. For any state $s$, if it satisfies $C_{\omega}(s,a^*)> C_{\omega}(s,a)$, where $a\in A\setminus\{a^*\}$. for any state $s$, we have
    \[
    V^{\pi_{\text{new}}} \geq V^{\pi}.
    \]
\end{theorem}

The proof is shown in Appendix~\ref{appendix_proof2}.
This assertion amplifies the significance of ensemble policies, showcasing the robustness of the integrated learning approach. By merging both policies, \ourmethod{} ensures that the newly derived policy does not compromise on quality.

\section{Experiments}
To demonstrate the impact of fine-grained state distinction on exploration and how memory reflection accelerates the convergence to optimal policies, we first conducted several visualization experiments in subsection~\ref{visualization}, verifying these components' individual and combined effects. As \ourmethod{} is flexible and can be easily integrated into existing algorithms, in the main experiment in subsection~\ref{mainexp}, we rigorously evaluate its performance in Montezuma's Revenge, one of the most challenging environments in the field. Additionally, in subsection~\ref{generalization}, we conducted extended experiments across a total of 56 environments, including the commonly used reinforcement learning platforms Atari and MuJoCo that cover scenarios ranging from dense to sparse reward environments and from continuous to discrete action spaces. These experiments confirmed the effectiveness and strong generalization of the \ourmethod{} framework.

We selected the integration of PPO with \ourmethod{} as the representative version of our method for large-scale comparisons across various environments. For the remainder of this paper, unless otherwise specified, \ourmethod{} refers to the integrated version of PPO and \ourmethod{}, and experiments are conducted with 50 million interaction steps.

\subsection{Visualization Analysis}
\label{visualization}
We selected the Cliff Walking, Four Rooms, and Dark Chamber environments for detailed visualization experiments and analytical argumentation. The settings of these environments are provided in Appendix~\ref{keshihua}. In each visualization experiment, we enable either the exploration or exploitation part of \ourmethod{} and disable the other part for a clear comparison.

\begin{figure*}[t]
    \centering
    \includegraphics[width=1.0\linewidth]{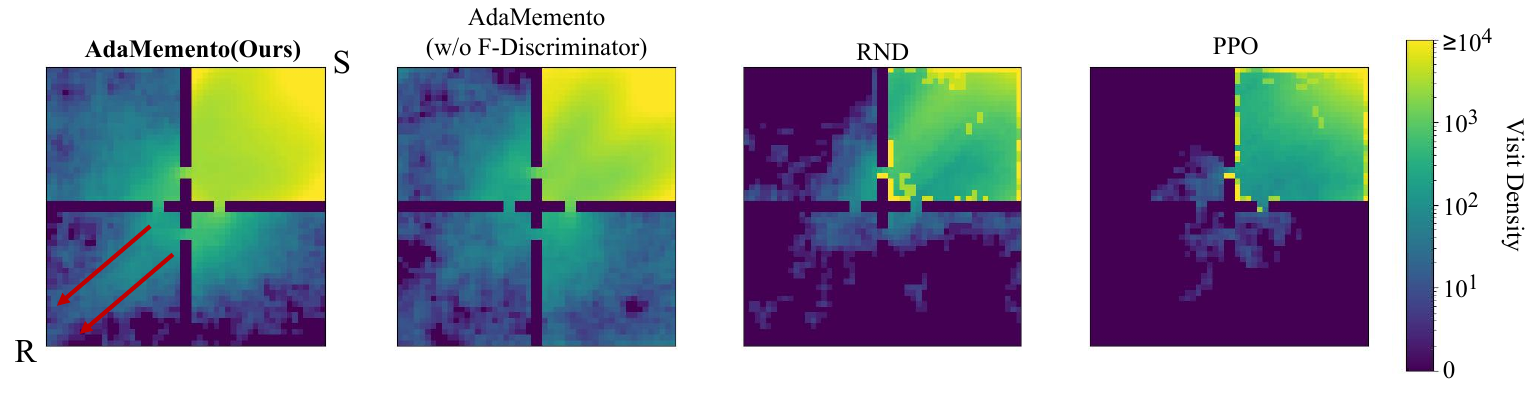}
    \caption{\textbf{Visualization Experiments in Four Rooms} in 0.5 million steps. \ourmethod{} can form an efficient strategy heading for the reward. }
    \label{fourroom}
\end{figure*}

\subsubsection{Visualization of the fine-grained distinction}
\paragraph{Dark Chamber.} 
The results of Dark Chamber are as shown in Figure~\ref{darkcharmber}, which indicate that both \ourmethod{} and \ourmethod{} (w/o F-Discriminator) outperform the baseline algorithms in terms of exploration range and depth. Moreover, \ourmethod{} with the state fine differentiation mechanism demonstrates superior deep exploration capabilities compared to its counterpart without this mechanism(the Ablation version). Comparing \ourmethod{} and \ourmethod{} (w/o F-Discriminator) with the domain's leading exploration policy RND, it is evident that generating reconstructed images using a single network to build intrinsic motivation is more effective than using two networks to generate feature vectors.

\paragraph{Four Rooms.} Figure~\ref{fourroom} indicates that \ourmethod{} achieves the largest exploration area and more comprehensive exploration of the environment, establishing a stable reward-acquiring policy, as illustrated by the red arrows. In contrast, the version without the state fine differentiation mechanism (the Ablation version), although superior to the baseline algorithms in exploration coverage, shows significantly less exploration area and quality than the full \ourmethod{}.

In summary, through comparative visualization experiments in both simple reward-free and complex sparse-reward environments, we conclude that our algorithm significantly outperforms the baseline algorithms. Moreover, the state fine differentiation mechanism clearly enhances exploration quality and depth, as evidenced by the comparison between RL methods with and without \ourmethod{} in both environments.

\subsubsection{Visualization of memory-reflection}
\begin{figure*}[t]
    \centering
    \includegraphics[width=1.0\linewidth]{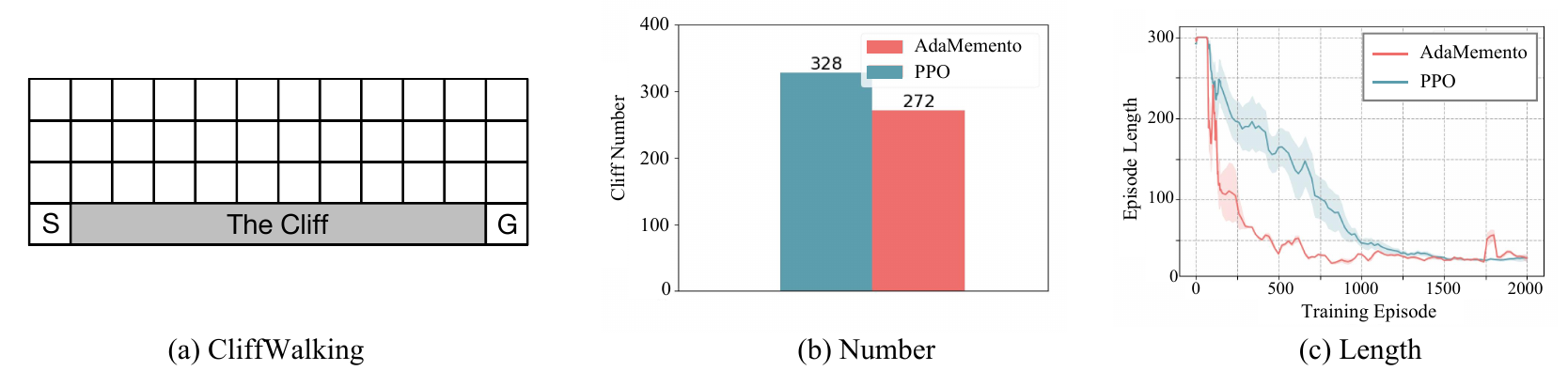}
    \caption{(a) is a Cliff Walking environment. (b) denotes the total number of times the agent falls off the cliff. (c) indicates the number of steps the agent takes to reach the goal G during training.}
    \label{cliff}
\end{figure*}

\paragraph{Cliff Walking.}
Figure~\ref{cliff}(b) shows that after integrating the memory-reflection module, the baseline significantly reduces the number of times it falls off the cliffs, indicating that the memory mechanism enables the agent to learn how to avoid dangers by reflecting on past failures. Figure~\ref{cliff}(c) demonstrates that the baseline model integrated with the memory reflection module reduces the number of steps from the starting point to the endpoint more quickly than the model without this module. This indicates that the memory reflection mechanism accelerates discovering the shortest path by summarizing the commonalities of successful experiences, thereby promoting policy learning and optimization and helping the model approach the optimal policy faster. This result validates the effectiveness of the memory reflection module in improving the efficiency of police training.

\subsection{Main experiments}
\label{mainexp}

We selected Montezuma's Revenge, one of the most challenging environments in RL, as the main experimental platform. For baselines, we compared our method with current state-of-the-art and widely used algorithms, including the classic PPO~\cite{schulman2017proximal}, the exploration-centric RND~\cite{burda2018exploration}, and the recently proposed RPO~\cite{reflectivepo}. Additionally, we compared mainstream memory utilization methods such as BT-NGU~\citep{campos2021beyond} for state storage, SIL and DTSIL for experience imitation and trajectory replay, and other memory-based methods.

Figure~\ref{montezumamain} shows the performance of the RL algorithms equipped with and without \ourmethod{}. The integrated algorithms outperform the baselines in the final score, convergence speed, and stability. Notably, in the case of PPO, integrating \ourmethod{} led to a score increase of 15 times, along with significant improvements in convergence speed and stability. Additionally, Figure~\ref{montezumamain}(b) compares \ourmethod{} with other memory baselines, demonstrating its superiority. This is due to our method's capability of summarizing shared knowledge from past experiences for policy prediction rather than simply imitating past behaviors, which promotes the discovery of new policies and leads to superior performance compared to other memory methods.

\subsection{Generalization experiments}
\label{generalization}

To demonstrate the broad applicability of \ourmethod{}, this subsection presents its performance in several well-known RL environments, including the remaining 55 discrete-space Atari games and the continuous-space environments of MuJoCo.

Figure~\ref{atariothers} shows the performance of \ourmethod{} in six challenging Atari environments~\citep{bellemare2016unifying}, excluding Montezuma. Except for a slight improvement in Venture, \ourmethod{} significantly improves performance in the other five environments. These results indicate that \ourmethod{} is effective in challenging discrete-space tasks. Additional results for the remaining 46 Atari environments are provided in Appendix~\ref{supplementary_experiment}.

Figure~\ref{mujucoothers} shows the performance of \ourmethod{} in continuous action spaces. The results indicate that our algorithm significantly outperforms the baseline in Swimmer, HalfCheetah, and HumanoidStandup environments. Notably, PPO integrated with \ourmethod{} achieved nearly three times the performance improvement in Swimmer and twice the improvement in HalfCheetah compared to the non-integrated version. These results demonstrate that \ourmethod{} is also effective in continuous spaces.

\section{Conclusion}

This paper introduces the \ourmethod{} framework, inspired by the human ability to learn from past experiences and predict the future. \ourmethod{} innovatively employs a memory reflection mechanism to guide the agent's predicted trajectory while avoiding potential pitfalls, accelerating the convergence of advantageous policies. Additionally, a coarse-fine distinction mechanism is proposed to promote precise exploration, providing high-quality reflection data for memory construction. For the coordination of exploration and exploitation, we leverage ensemble learning to perform joint inference, achieving an adaptive dynamic balance between exploration and exploitation and providing proof of the effectiveness of ensemble learning. This paper introduces a new intrinsic motivation paradigm in the RL field and theoretically proves its validity, eliminating the need for environment-specific designs while offering outstanding capabilities in terms of generalization and ease of integration. Experimental results across a wide range of environments demonstrate that \ourmethod{} exhibits strong robustness and generalization. \ourmethod{} exhibits theoretical innovation and demonstrates considerable potential in practical applications.

\bibliography{citations}
\bibliographystyle{iclr2025_conference}

\clearpage
\appendix
\section{Additional related works}
\subsection{Sparse Reward Environments}

Sparse reward environments~\citep{hare2019dealing,devidze2022exploration} in RL are scenarios where agents receive little to no feedback most of the time, only receiving rewards under specific conditions. This scarcity makes it difficult for agents to get enough guidance during training, causing traditional algorithms to perform poorly, as they rely on frequent reward signals to update policies and value functions.
To address this issue, researchers have proposed several approaches. First, improving exploration policies, such as using intrinsic rewards based on state novelty or prediction error, can encourage agents to explore unknown areas. Second, imitation and inverse reinforcement learning are widely applied, allowing agents to learn from expert demonstrations and find optimal paths more efficiently. Additionally, memory-based methods replay and mimic past successful experiences, substituting sparse reward signals, accelerating policy optimization, and easing the difficulty of updates in sparse reward environments.

\subsection{Memory motheds}

Existing mainstream memory utilization methods can be broadly divided into memory storage and experience reuse. Below, we provide a more detailed overview of related work to help readers gain a better understanding.

\paragraph{Storage-based methods:}

Memory storage is the simplest and most intuitive method of memory utilization. It helps address the sparse reward problem by storing the agent's experienced states to measure state novelty. A typical example of this approach is the NGU~\citep{badia2020never} method, which aims to solve exploration challenges in sparse reward environments by storing past state experiences. NGU memorizes processed state experiences and uses a clustering-based method to manage the memory buffer. By clustering visited states and using these cluster centers, the agent can identify and explore new states, providing additional intrinsic rewards based on state rarity. The agent receives higher intrinsic rewards when visiting rare states, encouraging the exploration of new areas. This memory comparison mechanism improves the agent's exploration efficiency and learning ability in sparse reward environments.
However, the NGU algorithm has some limitations. First, managing and updating the cluster centers can become increasingly complex in high-dimensional state spaces, reducing computational efficiency. Additionally, clustering-based methods may incorrectly group similar states with different values in high-dimensional spaces, overlooking critical nuances and affecting policy optimization. Finally, memory storage methods mainly focus on constructing state counts via memory buffers rather than using stored experiences to predict and guide future policies.

\paragraph{Experience Replay methods:}

Imitation learning~\citep{oh2018self,guo2020memory} based on experience replay is a memory utilization approach that centers on using the agent's past successful experiences to guide current policy optimization. Specifically, as the agent interacts with the environment, it stores information about its experienced states, actions, and corresponding rewards in a replay buffer. By analyzing this historical experience, particularly those associated with high rewards, the agent learns to imitate successful behaviors, reinforcing the positive effects of these experiences during policy optimization. Through replaying and mimicking high-reward actions, the agent improves decision-making quality, reduces reliance on random exploration, and accelerates policy convergence and improvement. By effectively leveraging valuable past experiences, such algorithms provide a more stable and efficient optimization path, helping the agent quickly learn and adapt to complex environments.
Despite significant progress in imitation learning based on experience replay, this approach has three notable limitations. First, these methods place too much emphasis on reproducing past high-value behaviors while falling short in exploring new policies. This can weaken the agent's exploration capabilities, causing it to miss the global optimal policy and become stuck in local optima. Second, these methods primarily focus on leveraging successful experiences while neglecting the value of failed experiences. Failed experiences are equally important for policy optimization. Human learning can extract valuable insights from both positive and negative experiences. In reinforcement learning, paths with high rewards often hide potential risks or pitfalls. Thus, effectively using failed experiences to avoid repeated mistakes is crucial for improving policy optimization. Third, these methods focus on imitating past successful behaviors rather than truly understanding the experiences and using them to predict future decisions. As a result, the agent often merely replicates past behavior patterns rather than using those experiences to anticipate future scenarios and outcomes. 

\subsection{Intrinsic Motivation}
In addressing the problem of sparse and delayed rewards, intrinsic motivation models~\cite{ryan2000intrinsic,badia2020agent57,oudeyer2007intrinsic} have gained increasing attention for their ability to enhance exploration efficiency. Intrinsic motivation methods simulate intrinsic drives in organisms, such as curiosity, to guide agents in exploration, thereby sustaining exploration efforts in environments lacking clear extrinsic rewards. These approaches can be broadly categorized into two types: count-based policies and prediction error-based policies.

\paragraph{Count-based Intrinsic Motivation:}

Count-based policies~\cite{bellemare2016unifying} track the frequency of an agent's state visits, using a decreasing function of state counts as exploration rewards. The basic principle is that the agent should receive higher rewards for visiting less frequently explored states, encouraging exploration of under-explored areas. The simplicity and intuitiveness of this mechanism have made it widely applicable in simple reinforcement learning environments. However, as the complexity of the environment increases, particularly in high-dimensional state spaces, the number of states grows rapidly. In such cases, the cost of maintaining and updating state counts significantly rises. Specifically, state counting requires substantial memory to record the visitation frequency of each state. Additionally, in dynamic environments, frequently updating the counts may lead to decreased computational efficiency. This not only increases resource consumption but may also slow down the agent's real-time decision-making, reducing its practicality in complex environments.
 
\paragraph{Error-based Intrinsic Motivation:}
Prediction error-based policies~\cite{burda2018exploration,pathak2017curiosity,burda2018large,achiam2017surprise,silvia2012curiosity} leverage neural network errors to assess state visit density, forming curiosity-driven exploration policies. Specifically, these methods calculate the agent's prediction error in a given state to estimate how frequently the state has been visited. Visits to low-density states are encouraged, while visits to high-density states are reduced, promoting exploration of unknown or novel states. This approach eliminates the dependence on storage and computational efficiency inherent in count-based intrinsic reward methods. It significantly improves exploration quality in the early stages of a task, allowing the agent to gather more information and experience. By this mechanism, the agent can more flexibly address exploration challenges in complex environments, actively seeking under-explored states.
This policy's effectiveness lies in broadening exploration and enabling the agent to adapt to environmental changes, thereby improving learning efficiency quickly. However, existing intrinsic motivation models often focus on enhancing exploration quality while paying less attention to exploitation. This tendency can lead to diminishing exploration returns in later stages as the agent's interaction with the environment deepens, causing exploration efficiency to decline significantly—a phenomenon known as the "intrinsic motivation fading problem"~\cite{barto2013novelty,oudeyer2007intrinsic}. Additionally, current prediction error-based intrinsic motivation methods tend to overlook subtle differences between similar states, which can result in missed opportunities to explore critical states and, consequently, potential policy improvements.

\section{Additional Preliminaries}
\subsection{Markov Decision Process (MDP)}
A Markov Decision Process (MDP)~\cite{puterman1990markov}, as described by Sutton and Barto (1999) and Puterman (2014), is defined by the tuple \( \langle S, A, P, R, \gamma \rangle \), where \( S \) and \( A \) denote the sets of states and actions, respectively. The function \( P \) represents the state transition probabilities, describing the likelihood of transitioning from one state to another upon taking a specific action. The reward function \( R \) assigns a scalar value (often assumed to be non-negative) for each state-action pair, reflecting the immediate feedback received after the transition.

The discount factor \( \gamma \in [0, 1) \) accounts for the agent’s preference for immediate versus future rewards by adjusting the value of rewards over time. During interactions in the MDP, the agent generates sequences (trajectories) consisting of states, actions, and rewards: \( \tau = (s_0, a_0, R(s_0, a_0)), \cdots, (s_t, a_t, R(s_t, a_t)), \cdots \). The value of a particular state-action pair under policy \( \pi \), denoted as the state-action value function \( Q^\pi(s_t, a_t) \), is calculated as follows:

\[
Q^\pi(s_t, a_t) = \mathbb{E}_{\tau \sim \pi} \left[ \sum_{i=0}^{\infty} \gamma^i R(s_{t+i}, a_{t+i}) \mid s_t, a_t \right].
\]

The agent’s behavior is defined by a policy, which is a probability distribution over actions given states. The policy \( \pi(a_i \mid s) \) in this context is derived using a softmax function applied to the state-action value function:

\[
\pi(a_i \mid s) = \frac{e^{Q(s, a_i)}}{\sum_k e^{Q(s, a_k)}}, \quad i = 1, 2, \cdots, n.
\]

To ensure that the agent chooses actions that maximize its long-term accumulated rewards, the aim of policy optimization is to derive a policy \( \pi \) that maximizes the expected cumulative discounted reward \( \eta(\pi) \). This is given by:

\[
\eta(\pi) = \mathbb{E}_{\tau \sim \pi} \left[ \sum_{i=0}^{\infty} \gamma^i R(s_i, a_i) \right].
\]

The objective is to encourage the agent to make decisions that yield the highest total reward, balancing both immediate and future returns.

\subsection{Intrinsic-Based Reinforcement Learning}
Extending the traditional Markov Decision Process (MDP) framework, intrinsic rewards are introduced to facilitate agent exploration, particularly in environments characterized by sparse extrinsic rewards. In such settings, relying solely on extrinsic feedback can lead to inefficient exploration and poor learning outcomes, as the agent may not receive enough signals to optimize its policy effectively.

Intrinsic rewards serve as an additional incentive for the agent, promoting the discovery of novel states and encouraging behaviors that might eventually lead to higher extrinsic rewards. The total reward function \( R_{\text{total}}(s, a) \) is thus composed of both extrinsic rewards \( R_{\text{ext}} \) and intrinsic rewards \( R_{\text{int}} \), which are combined to form a more comprehensive feedback mechanism:

\[
R_{\text{total}}(s, a) = R_{\text{ext}}(s, a) + \alpha R_{\text{int}}(s, a).
\]

The intrinsic reward \( R_{\text{int}}(s, a) \) often reflects properties such as state novelty, prediction error, or curiosity. These intrinsic signals help agents maintain a high level of exploration, ensuring that less frequently visited states are considered, thus improving the learning of optimal policies over time. By incorporating intrinsic rewards, intrinsic-based reinforcement learning aims to balance exploration and exploitation, allowing agents to learn effectively even when extrinsic rewards are sparse or delayed. This approach significantly improves learning efficiency in complex environments where traditional methods struggle to provide sufficient guidance.

\section{Additional Proofs}

\subsection{Proof of Theorem~\ref{the_same_ap1}}
\label{appendix_proof1}

\begin{proof}
    Under Assumption \ref{assu1}, we define $C \triangleq (1-\gamma)\min_{s}[Q^*(s, a^*)-Q^*(s,a^{sub})]$. So, for any state $s$ and action $a$, we omit ($s,a$), and have
    \begin{equation*}
        R \leq R +\beta R_{I}\leq R+C
    \end{equation*}
    Then 
    \begin{equation} \label{equation_1}
	Q^*(s,a) \triangleq \max_{\pi}\mathbb{E}_{\tau}\left[\sum_t \gamma^t R_t|s,a\right]
	\leq \max_{\pi}\mathbb{E}_{\tau}\left[\sum_t \gamma^t (R_t+\beta R_I)|s,a\right] \triangleq Q_1^*(s,a)
    \end{equation}
    and
    \begin{equation}\label{equation_2}
	Q_1^*(s,a) = \max_{\pi}\mathbb{E}_{\tau}\left[\sum_t \gamma^t (R_t+\beta R_I)|s,a\right]
	\leq \max_{\pi}\mathbb{E}_{\tau}\left[\sum_t \gamma^t (R_t+C)|s,a\right]
	\triangleq \hat{Q}^*(s,a)
    \end{equation}
    According to the paper \citep{Sal}, we have 
    \begin{equation}\label{equation_3}
        \hat{Q}^*(s,a)\triangleq\max_{\pi} \mathbb{E}_{\tau}\left[ \sum_{t}\gamma^t (R_t + C)|s,a\right]= Q^*(s,a)+\frac{C}{1-\gamma}
    \end{equation}
    where $Q^*$ is the optimal $Q$ function.
    Since $a^*$ and $a^{sub}$ are optimal and suboptimal actions under the state $s$, respectively, we have
    $Q^*(s, a^*)> Q^*(s,a^{sub})$. Next, we will prove that $Q_1^*(s, a^*)> Q_1^*(s,a^{sub})$ holds.

    For the optimal action $a^*$ and suboptimal action $a^{sub}$, we have
    \begin{align*}
	& Q^*(s,a^*)\leq Q_1^*(s,a^*) \leq \hat{Q}^*(s,a^*)=Q^*(s,a^*)+\frac{C}{1-\gamma}\\
	& Q^*(s,a^{sub})\leq Q_1^*(s,a^{sub}) \leq \hat{Q}^*(s,a^{sub})=Q^*(s,a^{sub})+\frac{C}{1-\gamma}
    \end{align*}
    By simplifying, we have
    \begin{align*}
	 & Q_1^*(s,a^*)- Q_1^*(s,a^{sub})
	 > Q^*(s,a^*)-Q^*(s,a^{sub})-\frac{C}{1-\gamma}
    \end{align*}
    By definition of $C$, we know $Q^*(s,a^*)-Q^*(s,a^{sub})-\frac{C}{1-\gamma}>0$ holds.
    Therefore, $Q_1^*(s, a^*)> Q_1^*(s,a^{sub})$ holds.
\end{proof}

\subsection{Proof of Theorem~\ref{the_same_ap2}}
\label{appendix_proof2}

\begin{proof}
    Since the value of $C_{\omega}(s, a)$ is related to the state $s$ and action $a$. Under state $s$, due to $C_{\omega}(s,a^*)> C_{\omega}(s,a)$, we can get 
    \begin{equation*}
    \begin{aligned}
        &\sum_{a} I_{\{C_{\omega}(s,a)\geq \kappa\}} \pi(a|s)Q^{\pi}(s,a)\\
        \leq & I_{\{C_{\omega}(s,a^*)\geq \kappa\}} \sum_{a} \pi(a|s)Q^{\pi}(s,a)\\
        \leq&  I_{\{C_{\omega}(s,a^*)\geq \kappa\}} Q^{\pi}(s,a^*)\\
        =   &\sum_{a} I_{\{C_{\omega}(s,a)\geq \kappa\}} \pi_\theta(a|s)Q^{\pi}(s,a)
    \end{aligned}
    \end{equation*}
    Therefore, we get
    \begin{align*} 
		 V^{\pi}(s) 
		& = \sum_{a}\pi(a|s) Q^{\pi}(s,a)\\
            & = \sum_{a} \left(1-I_{\{C_{\omega}(s,a)\geq \kappa\}}\right) \pi(a|s)Q^{\pi}(s,a) + \sum_{a} I_{\{C_{\omega}(s,a)\geq \kappa\}} \pi(a|s)Q^{\pi}(s,a)\\
		& \leq \sum_{a} \left(1-I_{\{C_{\omega}(s,a)\geq \kappa\}}\right) \pi(a|s)Q^{\pi}(s,a) + \sum_{a} I_{\{C_{\omega}(s,a)\geq \kappa\}} \pi^*(a|s)Q^{\pi}(s,a)\\
		& = \sum_{a}\pi_{\text{new}}(a|s) Q^{\pi}(s)\\
		& = E_{\pi_{\text{new}}} \{ Q^{\pi}(s, a)| s_{t}=s\}\\
		& = E_{\pi_{\text{new}}} \{r_t + \gamma V^{\pi}(s_{t+1})| s_{t}=s\}\\
		& \leq E_{\pi_{\text{new}}} \{r_t + \gamma \sum_{a_{t+1}}\pi_{\text{new}}(a_{t+1}|s_{t+1})Q^{\pi}(s_{t+1})| s_{t}=s\}\\
		& \vdots\\
		& \leq V^{\pi_{\text{new}}}(s).
	\end{align*} 

\end{proof}

\section{Supplementary Experiment Details}
\label{supplementary_experiment}

\subsection{Construction and update of memory buffers}
\label{supplementary_buffers}
In memory-reflection module, maintain two fix-sized memory buffer of history trajectories, namely the memory buffer and the reflection buffer.

\paragraph{Memory Buffer:}

Memory mechanisms must not only adapt to new environments by incorporating new experiences but also ensure timely forgetting (or replacing) of outdated experiences to maintain a reasonable capacity and avoid excessive overhead. Based on this, we maintain a fixed-size buffer of advantageous trajectories, \( M = \{T_1, T_2, T_3, \dots, T_i\} \), where each \( T_i \) contains a complete sequence of state-action-reward tuples: 
\[
T_i = \{(s_1, a_1, r_1), (s_2, a_2, r_2), \dots, (R_i, N_i)\}.
\]
Here, \( s_i \), \( a_i \), and \( r_i \) represent the state, action, and reward, respectively, \( N_i \) is the total length of the trajectory excluding terminal (death) states, and \( R_i \) is the total reward accumulated in the trajectory. If the trajectory ends in a non-terminal state, it is entirely stored.

The update strategy of memory buffer works as follows: if \( R_i = R_k \), the trajectory with the shorter length, \( N_i \) or \( N_k \), replaces the longer one; if \( R_i < R_k \), the entire trajectory \( T_i \) is replaced by \( T_k \). The buffer \( G \) is used for training both the Prediction Network and the Reflection Network.

\paragraph{Reflection Buffer:}

In addition to utilizing successful experiences to optimize policies, it is crucial to repeatedly reflect on failed experiences until the agent learns to avoid potential risks, thereby creating more opportunities for deeper exploration. Therefore, unlike existing memory mechanisms focusing solely on advantageous experiences, our approach also emphasizes learning from failures. The buffer $R$ maintains a certain amount of failed prediction experiences. These experiences result from the Reflection Network's incorrect assessments of predicted actions. The failed experiences are repeatedly fed into the Reflection Network for training until the predictions are accurate, and the agent learns to avoid dangers.

\subsection{Visualization of Experimental Environments Setup}
\label{keshihua}
\begin{figure*}[t]
    \centering
    \includegraphics[width=0.8\linewidth]{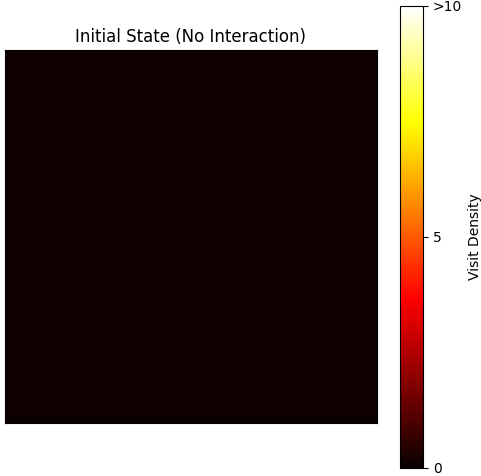}
    \caption{\textbf{Dark Chamber environment} This environment contains no rewards or traps and is designed as a testing platform to evaluate the algorithm's exploration capabilities under reward-free conditions.}
    \label{darkcharmberenv}
\end{figure*}

\paragraph{Dark Chamber.} To demonstrate the exploration capabilities of various algorithms in a reward-free environment, we constructed the \emph{Dark Chamber} environment, as shown in Figure ~\ref{darkcharmberenv}. This environment is 50x50 in size, with no rewards or penalties. The agent starts exploring from the bottom left, where unexplored areas are represented as black fog, and the color of explored areas gradually lightens based on exploration density.

\begin{figure*}[t]
    \centering
    \includegraphics[width=0.8\linewidth]{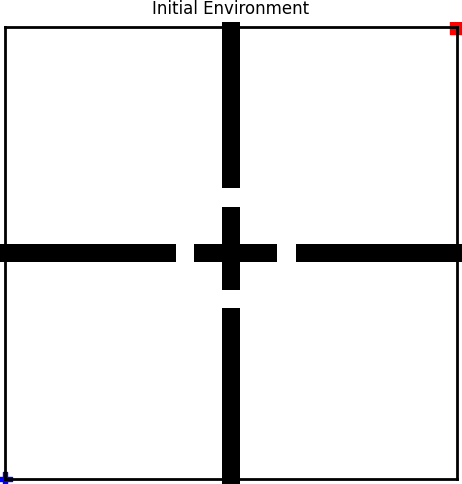}
    \caption{\textbf{Four Rooms environment} The environment features a central obstacle made of four walls, with only one door the same size as the agent for it to pass through. The agent starts at the red point in the top right corner and must navigate through the obstacle to reach the blue point in the bottom left corner to obtain the only reward. As a setting with sparse rewards and obstacles, this environment is ideal for evaluating an algorithm's ability to explore challenging scenarios.}
    \label{fourroomenv}
\end{figure*}

\paragraph{Four Rooms.} To further evaluate the agent's exploration ability in a complex environment with sparse rewards and obstacles, we selected the \emph{Four Rooms} environment for comparative visualization experiments, as shown in Figure~\ref{fourroomenv}. The agent starts from point S in the upper right, explores the environment, avoids the cross-shaped wall in the middle, and passes through the small doorway to reach the goal in the lower left for a reward.

\paragraph{Cliff Walking.}
The Cliff Walking environment, filled with cliff traps, was selected as the visualization platform to evaluate the memory-reflection mechanism, as shown in Figure~\ref{cliff}(a). In this environment, the gray areas represent cliffs, and the agent must move from point S to point G. To succeed, the agent needs to learn to avoid dangers while reaching goal G using a relatively short path, making this environment highly suitable for verifying the effectiveness of the memory-reflection mechanism.

\subsection{Algorithm Description}
The algorithmic process of AdaMemento can be seen in Algorithm ~\ref{alg:adamemento}.
\begin{algorithm}
\caption{AdaMemento: Adaptive Memory-Assisted Policy Optimization}
\label{alg:adamemento}
\begin{algorithmic}[1]
\Require Environment $\mathcal{E}$, Policy Network Parameters $\theta$, Discount Factor $\gamma$, Batch Size $n$, Memory Buffer $M$, Reflection Buffer $R$, Intrinsic Reward Parameter $\beta$, Learning Rate $\alpha$
\State \textbf{Initialization:}
    \State Initialize policy network parameters $\theta$, memory buffer $M$, reflection buffer $R$, intrinsic reward parameter $\beta$, learning rate $\alpha$
    
\For{$t = 0, 1, 2, \dots$}
    \State \textbf{1. Data Collection:}
    \State Use current policy $\pi_{\theta}$ to interact with environment $\mathcal{E}$ and collect $n$ trajectories.
    
    \State \textbf{2. Memory Reflection Module:}
    \State Sample from memory buffer $M$ to update \textbf{Predictive Network}, summarizing shared experiences across trajectories.
    \State Sample from both memory buffer $M$ and reflection buffer $R$ to update \textbf{Reflection Network}, adjusting action confidence based on predictive error.

    \State \textbf{3. Coarse-fine distinction module:}
    \State Compute reward based on state reconstruction error and sparsity to encourage exploration of low-density states.
    \State Promote exploration of states with high novelty or insufficient exploration.

    \State \textbf{4. Exploration-Exploitation ensemble learning:}
    \State Based on the confidence level of the Reflection Network, adaptively select actions from memory, reflection, or stochastic policies.
    \State Update policy network parameters $\theta$ by the following rule:
    \[
    \theta \leftarrow \theta + \alpha \nabla_{\theta} \mathcal{L}(\pi_\theta)
    \]
    where $\mathcal{L}(\pi_\theta)$ is the optimization objective combining memory and intrinsic reward.
    
\EndFor
\end{algorithmic}
\end{algorithm}

\subsection{Parameter Settings}
In this subsection, we present the parameter settings for the primary experimental platform, Atari, used in the majority of this paper. It is important to note that these settings represent only one possible configuration for running AdaMemento and do not necessarily guarantee optimal performance. For detailed parameter settings, see Table ~\ref{tab:hyperparameters}.
\begin{table*}[h!]
    \centering
    \caption{Hyperparameter Table}
    \renewcommand{\arraystretch}{1.2} 
    \begin{tabular}{>{\centering\arraybackslash}p{0.5\textwidth} >{\centering\arraybackslash}p{0.5\textwidth}}
        \toprule
        Hyperparameter & Value \\
        \midrule
        MaxStepPerEpisode   & 4500 \\
        ExtCoef             & 2.0 \\
        LearningRate        & 1e-4 \\
        NumEnv              & 32 \\
        NumStep             & 128 \\
        Gamma               & 0.999 \\
        IntGamma            & 0.99 \\
        Lambda              & 0.95 \\
        StableEps           & 1e-8 \\
        StateStackSize      & 4 \\
        PreProcHeight       & 84 \\
        ProProcWidth        & 84 \\
        UseGAE              & True \\
        UseGPU              & True \\
        UseNorm             & False \\
        UseNoisyNet         & False \\
        ClipGradNorm        & 0.5 \\
        Entropy             & 0.001 \\
        Epoch               & 4 \\
        MiniBatch           & 4 \\
        PPOEps              & 0.1 \\
        IntCoef             & 1.0 \\
        StickyAction        & True \\
        ActionProb          & 0.25 \\
        UpdateProportion    & 0.25 \\
        ObsNormStep         & 50 \\
        Confidence          & 0.85 \\
        GoodBufferSize      & 10 \\
        BadBufferSize       & 5000 \\
        GoodBufferBatchSize & 1 \\
        BadBufferBatchSize  & 128 \\
        CommonBufferSize    & 100000 \\
        OriPolicyEnvNum     & 16 \\
        ExploitUpdate       & 50 \\
        \bottomrule
    \end{tabular}
    \label{tab:hyperparameters}
\end{table*}

\section{\ourmethod{} vs. Baseline Performance in All Atari Environments}
\begin{figure*}
    \centering
    \includegraphics[width=0.9\linewidth]{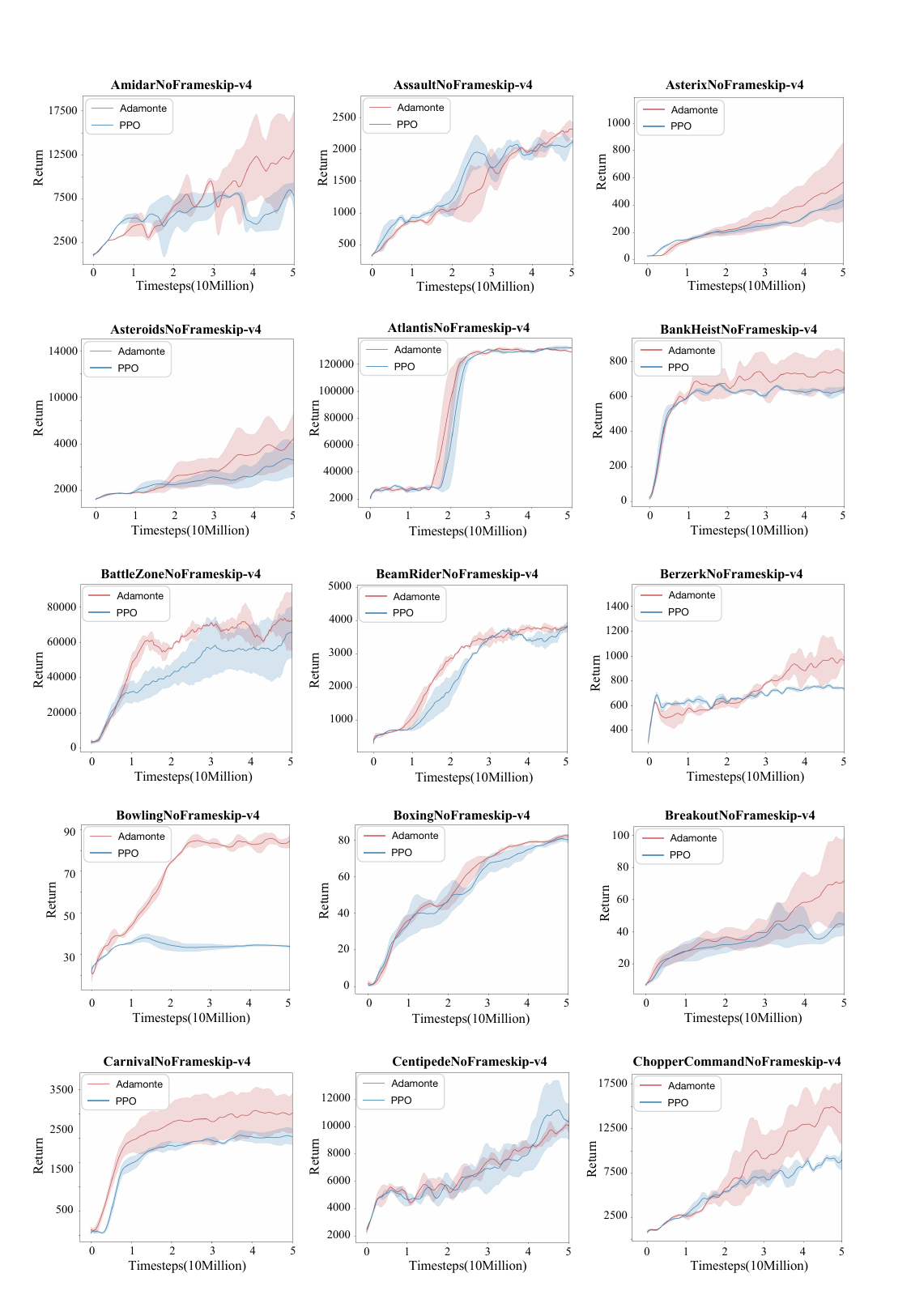}
    \phantomcaption
\end{figure*}
\begin{figure*}
    \centering
    \includegraphics[width=0.9\linewidth]{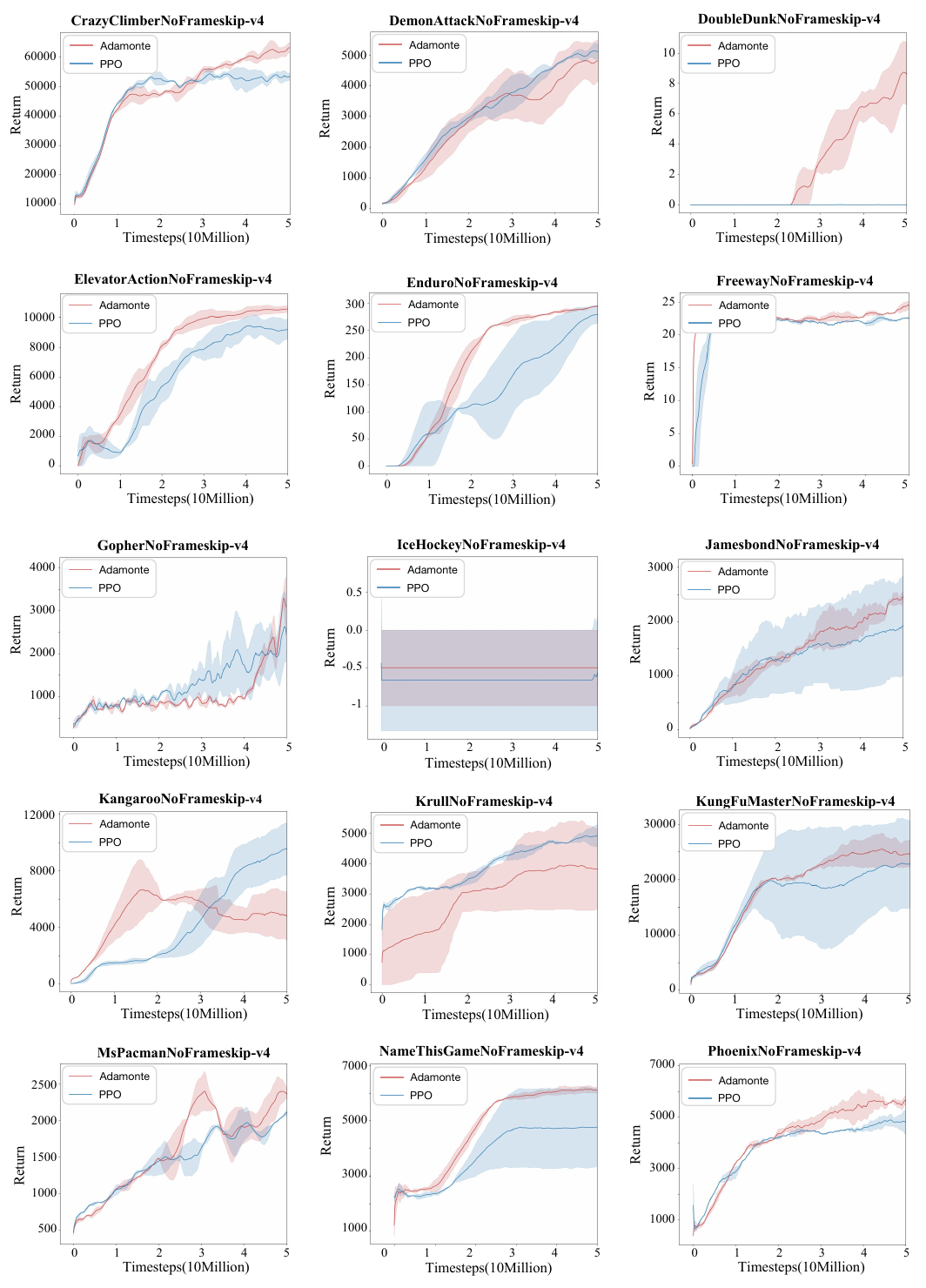}
    \phantomcaption
\end{figure*}
\begin{figure*}
    \centering
    \includegraphics[width=0.9\linewidth]{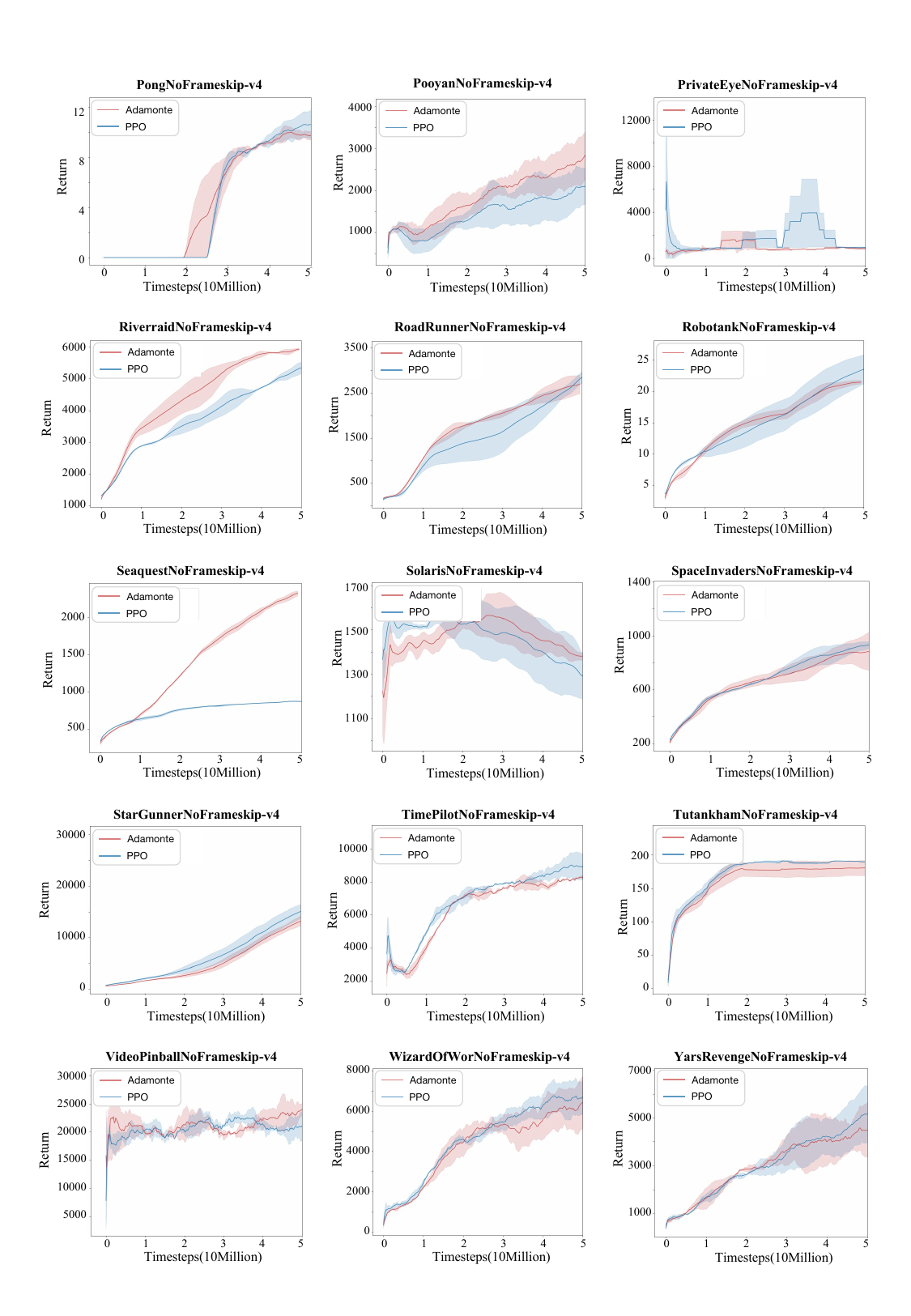}
    \phantomcaption
\end{figure*}
\begin{figure*}
    \centering
    \includegraphics[width=0.9\linewidth]{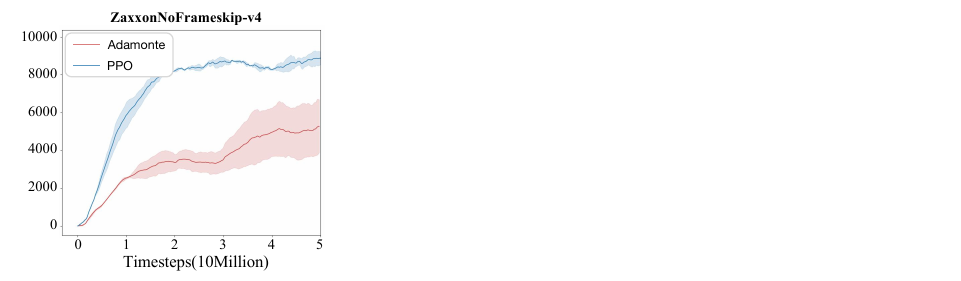}
    \caption{\textbf{Supplementary experiments in Atari.} 
    }
    \label{atari}
\end{figure*}

\end{CJK*}

\end{document}